\documentclass[conference]{IEEEtran}
\IEEEoverridecommandlockouts
\usepackage{cite}
\usepackage{amsmath,amssymb,amsfonts}
\usepackage[ruled,vlined,linesnumbered,noend]{algorithm2e}
\usepackage{graphicx}
\usepackage{textcomp}
\usepackage{color}
\usepackage{xcolor}
\usepackage{amsmath}
\usepackage{siunitx}
\usepackage{mathtools}
\usepackage{breqn}
\usepackage{multirow}
\usepackage{makecell}
\def\BibTeX{{\rm B\kern-.05em{\sc i\kern-.025em b}\kern-.08em
    T\kern-.1667em\lower.7ex\hbox{E}\kern-.125emX}}

\usepackage{cleveref}

\begin{document}

\title{Path-Tracking Hybrid A* and Hierarchical MPC Framework for Autonomous Agricultural Vehicles
}

\author{\centering\IEEEauthorblockN{1\textsuperscript{st} Mingke Lu}
\IEEEauthorblockA{\textit{College of Engineering} \\
\textit{Peking University}\\
Beijing, China \\
lmk@stu.pku.edu.cn}
\and
\IEEEauthorblockN{2\textsuperscript{nd} Han Gao}
\IEEEauthorblockA{\textit{College of Engineering} \\
\textit{Peking University}\\
Beijing, China \\
hangaocoe@pku.edu.cn}
\and
\IEEEauthorblockN{3\textsuperscript{rd} Haijie Dai}
\IEEEauthorblockA{\textit{College of Engineering} \\
\textit{Peking University}\\
Beijing, China \\
2401111763@stu.pku.edu.cn}
\and
\IEEEauthorblockN{4\textsuperscript{th} Qianli Lei}
\IEEEauthorblockA{\textit{College of Engineering} \\
\textit{Peking University}\\
Beijing, China \\
2401112013@stu.pku.edu.cn}
\and
\IEEEauthorblockN{5\textsuperscript{th} Chang Liu}
\IEEEauthorblockA{\textit{College of Engineering} \\
\textit{Peking University}\\
Beijing, China \\
changliucoe@pku.edu.cn}
}
\maketitle
\begin{abstract}
We propose a Path-Tracking Hybrid A* planner coupled with a hierarchical Model Predictive Control (MPC) framework for path smoothing in agricultural vehicles. The goal is to minimize deviation from reference paths during cross-furrow operations, thereby optimizing operational efficiency, preventing crop and soil damage, while also enforcing curvature constraints and ensuring full-body collision avoidance. Our contributions are threefold: (1) We develop the Path-Tracking Hybrid A* algorithm to generate smooth trajectories that closely adhere to the reference trajectory, respect strict curvature constraints, and satisfy full-body collision avoidance. The adherence is achieved by designing novel cost and heuristic functions to minimize tracking errors under nonholonomic constraints. (2) We introduce an online replanning strategy as an extension that enables real-time avoidance of unforeseen obstacles, while leveraging pruning techniques to enhance computational efficiency.  (3) We design a hierarchical MPC framework that ensures tight path adherence and real-time satisfaction of vehicle constraints, including nonholonomic dynamics and full-body collision avoidance. By using linearized MPC to warm-start the nonlinear solver, the framework improves the convergence of nonlinear optimization with minimal loss in accuracy. Simulations on real-world farm datasets demonstrate superior performance compared to baseline methods in safety, path adherence, computation speed, and real-time obstacle avoidance.
\end{abstract}


\section{Introduction}
With the rapid advancement of robotics technologies \cite{wang2022applications}, 
\cite{cheng2023recent}, autonomous mobile robots have demonstrated great potential in various agricultural applications, including crop monitoring \cite{bayati2018mobile}, precision spraying \cite{meshram2022pesticide}, harvesting \cite{fue2020extensive}, and autonomous weeding \cite{bawden2017robot}. These robots enable high-efficiency operations in large-scale farmlands, significantly reducing human labor and improving productivity.

In this context, path planning emerges as a fundamental capability for agricultural robots. Prior studies have explored diverse challenges, such as terrain irregularity—addressed through attitude stabilization and center-of-mass constraints \cite{santos2020path1}, multi-task execution—such as simultaneous monitoring and navigation during cultivation \cite{mai2019path}, and field coverage—optimized via systematic sweeping and coverage strategies \cite{wang2025hybrid},
\cite{lei2022deep},
\cite{din2022deep}. Additional works also consider energy-efficient routing \cite{yan2020agrirover}, among other practical issues.

An important yet insufficiently studied task in agricultural robotics is path planning for cross-furrow operations, where vehicles must traverse furrows while closely following a predefined reference trajectory, typically obtained through manual annotation or front-end planning algorithms. The primary objective in this scenario is to minimize deviation from the reference path, as even small lateral errors may lead to crop damage, soil compaction, or inefficiencies in field operations.

Path smoothing techniques—which refine raw paths into smooth and kinematically feasible trajectories—have shown promise in addressing complex navigation tasks. For example, interpolation-based methods using Bézier or B-spline curves \cite{yang2010analytical, noreen2020collision}, special-curve approaches such as clothoids and Dubins paths \cite{lambert2021optimal, reeds1990optimal}, and optimization-based methods \cite{song2021improved, feng2022path} have been widely applied in general-purpose robotics.  

However, these methods face significant limitations in cross-furrow agricultural scenarios:
First, they often prioritize geometric smoothness or curvature constraints, but rarely aim to minimize deviation from a reference trajectory, which is crucial to prevent crop damage.
Second, most approaches assume point-mass models, overlooking the vehicle’s full dimensions—making them unsuitable in narrow farmland settings where whole-body collision avoidance is required. Therefore, generating kinematically feasible trajectories that closely follow reference paths in constrained farmland remains an open challenge.

To bridge this gap, we propose a novel cross-furrow path planning and control framework for agricultural vehicles with three key contributions. 
First, we develop the Path-Tracking Hybrid A* algorithm to generate smooth trajectories that adhere closely to the reference trajectory while respecting strict curvature constraints and satisfying full-body collision avoidance. The adherence is achieved by designing novel cost and heuristic functions to minimize tracking errors under nonholonomic constraints.
Second, we propose an online replanning strategy as an extension of Path-Tracking Hybrid A* that enables real-time avoidance of unforeseen obstacles, while leveraging pruning techniques to enhance computational efficiency. 
Third, we design a hierarchical Model Predictive Control (MPC) framework that ensures tight path adherence and real-time satisfaction of vehicle constraints, including nonholonomic dynamics and full-body collision avoidance. By using linearized MPC to warm-start the nonlinear solver, the framework improves convergence of the nonlinear optimization with a minor loss in accuracy.
Simulations on real-world farm datasets demonstrate superior performance over baselines in safety, adherence to reference paths, computation speed, and real-time response to unforeseen obstacles.

The structure of this paper is organized as follows. In \Cref{formulation}, the model for the farm and agricultural vehicles is formulated. \Cref{deviation} and \Cref{path tracking} describe the proposed Path-Tracking Hybrid A* approach, with a focus on the deviation-aware cost and heuristic functions. \Cref{subsubsec:pruning} introduces the pruning techniques to enhance computational speed, and \cref{real-time-theory} presents the real-time extension of the Path-Tracking Hybrid A* algorithm, designed to handle unforeseen obstacles. \Cref{MPC} outlines the hierarchical MPC strategy utilized for trajectory tracking. Finally, \Cref{simulation} provides comparative simulation results against baseline methods, covering both static path tracking and real-time obstacle avoidance scenarios.

\section{Related Work}
\subsection{Hybrid A* in Path Planning}
In path planning, the process is typically divided into a front-end and a back-end. The front-end focuses on generating a raw path from the start to the goal while avoiding obstacles, often using sampling- or search-based algorithms such as Hybrid A* \cite{dolgov2008practical}.
The back-end, or path smoothing, refines this raw path by considering the vehicle's kinematic constraints and other factors, ensuring the trajectory is feasible and smooth for the robot to follow.

In recent years, a substantial body of research has explored the Hybrid A* algorithm and its variants for path planning applications \cite{sedighi2019guided, sheng2021autonomous, zheng2023navigation, chang2024hybrid, meng2023improved}. Hybrid A* integrates the discrete grid-based search of A* with continuous-space refinement, enabling the generation of kinematically feasible and smooth trajectories. Recent studies have enhanced the algorithm's performance by incorporating it with various techniques, such as visibility diagrams \cite{sedighi2019guided}, safe travel corridor models \cite{sheng2021autonomous}, time-elastic bands \cite{zheng2023navigation}, hierarchical clustering \cite{chang2024hybrid}, and multistage dynamic optimization frameworks \cite{meng2023improved}. These efforts primarily aim to improve computational efficiency and local trajectory quality. However, to the best of our knowledge, Hybrid A* has been predominantly employed as a front-end planner in navigation or parking contexts, and its potential for path smoothing has yet to be explored.

\subsection{Path Smoothing Techniques}
State-of-the-art path smoothing approaches can be broadly classified into three categories: interpolation-based, special-curve-based, and optimization-based methods \cite{ravankar2018path}.

Interpolation-based methods employ parameterized curves—such as polynomial, Bézier, and B-spline curves—to fit coarse paths generated by front-end planners. While polynomial interpolation is classical, it suffers from limitations like Runge’s phenomenon \cite{epperson1987runge}. Bézier curves are well-studied for enforcing $C^2$ continuity and curvature constraints \cite{yang2010analytical}. Recent advances use low-order Bézier curves (e.g., quadratic) to reduce control point complexity \cite{durakli2022new} and enable curvature-continuous path generation \cite{liu2024high}. Similarly, B-splines support real-time generation with smoothness and curvature constraints \cite{elbanhawi2015continuous}, and are often integrated with collision avoidance modules \cite{noreen2020collision}.

Special-curve-based methods utilize geometric primitives such as clothoids and Dubins curves, which are more difficult to parameterize than interpolation curves but naturally reflect vehicle kinematics. Clothoids, or Euler spirals, feature linearly varying curvature and have been applied in high-speed autonomous navigation \cite{lambert2021optimal} and emergency stopping path design \cite{lin2024clothoid}. Dubins and Reeds-Shepp curves \cite{dubins1957curves, reeds1990optimal} consist of circular arcs and straight lines, and are commonly used for generating shortest feasible paths under nonholonomic constraints \cite{park2022three}.

Optimization-based methods typically augment the above techniques by formulating the smoothing task as a multi-objective optimization problem. These methods can jointly consider collision avoidance, dynamic constraints (e.g., speed, acceleration, jerk), and environmental factors. Metaheuristic algorithms such as particle swarm optimization (PSO) have been applied to Bézier-based smoothing \cite{ song2021improved, xu2022new}, while ant colony optimization (ACO) has been used in B-spline smoothing frameworks \cite{feng2022path, huo2024new}.

However, these methods often prioritize geometric smoothness or curvature constraints, while neglecting to minimize deviation from a reference trajectory—an essential factor for crop protection. Additionally, their reliance on point-mass models fails to account for the vehicle’s full dimensions, limiting their applicability in constrained farmland environments where whole-body collision avoidance is critical.

\subsection{Path Planning in Agricultural Robotics}
The work on path planning for agricultural robots can be mainly divided into three areas: point-to-point planning, coverage planning, and multi-robot collaborative path planning, which will be introduced separately in the following. 

In Point-to-Point path planning, Voronoi diagrams \cite{habib2016mobile} and variants of Ant Colony Optimization (ACO) \cite{li2022path} are employed to generate safe and optimal paths from the start to the end point. Additionally, learning-based methods are also utilized; for instance, \cite{zhao2024improved} employs a 3D neural network to enable collision-free navigation for robots with arbitrary shapes, while \cite{yang2022intelligent} applies deep reinforcement learning to facilitate dynamic obstacle avoidance.

For complete coverage path planning (CCPP), the most commonly used basic methods for simple path planning are the nested and reciprocating approaches \cite{ning2022research}. Hybrid variants, which combine these basic methods with other techniques, demonstrate improved performance. Notable examples include the use of fully convolutional neural networks (FCNN) for workspace edge detection and optimized path generation \cite{lei2022deep}, inner spiral methods to reduce non-working path length \cite{wang2025hybrid}, and a CNN-based Dual Deep Q-learning (DDQN) algorithm that adapts to dynamic agricultural environments \cite{din2022deep}.

In the field of multi-robot collaborative path planning, several studies focus on simultaneous task and path planning to enhance coverage efficiency, employing methods such as the improved Dijkstra algorithm \cite{wang2023collaborative} and particle swarm optimization \cite{liu2021optimization}. Additionally, path conflicts between different robots are a critical concern, which is addressed in \cite{cao2023global} using topographic maps and time windows.  

 However, the aforementioned methods neither explicitly consider the adherence to reference trajectory tracking nor incorporate full-body collision avoidance within the planning framework.


\section{Problem Formulation}
\label{formulation}
The objective is to generate motion trajectories that closely follow a given reference path—either manually annotated or generated by a front-end algorithm—while satisfying both nonholonomic constraints and collision-avoidance constraints with respect to the field boundaries.

The farmland is formulated as a two-dimensional polygonal region as a two-dimensional polygon-shaped region $P\subset\mathbb{R}^2$, which is defined by the polygon vertices $\mathcal{V} = \{\boldsymbol{v}_1, \boldsymbol{v}_2, \dots, \boldsymbol{v}_n\}$ and edges $\mathcal{E} = \{e_1, e_2, \dots, e_n\}$.
Within the farm $P$, the movement of agricultural vehicles should align with the cross-furrow reference paths.  Each reference path \( R \subset P \) is represented as a polyline, defined by an ordered sequence of points \( \{\boldsymbol{r}_0, \boldsymbol{r}_1, \dots, \boldsymbol{r}_{m}\} \subset P \). The polyline consists of line segments \( \mathcal{L} = \{l_1, l_2, \dots, l_m\} \), where each segment \( l_j \) connects \( r_{j-1} \) and \( r_{j} \). Note that the start point is $\boldsymbol{r}_0$ and the end point is $\boldsymbol{r}_m$. The orientation of each segment $l_j$ is denoted as $\phi_j$.

The agricultural vehicle is formulated as a Wheeled Mobile Robot (WMR).
Specifically, the kinematics of the WMR is formulated as the following bicycle model:
\begin{equation}
    \begin{cases}
    \dot{x} = v \cos\theta, \\
    \dot{y} = v \sin\theta, \\
    \dot{\theta} = v\frac{tan\delta}{L},\\
    \dot{v} = a,
    \end{cases}
    \label{eq:case1}
\end{equation}
where $L$ denotes the wheelbase, and the state vector $\boldsymbol{x} = [x, y, \theta, v]^T$ indicates the configuration of the WMR, including the position $[x,y]^T\in\mathbb{R}^2$, orientation $\theta\in[0,2\pi]$, and linear velocity $v$ of the rear wheel axis center $C$.
The control input $\boldsymbol{u} = [\delta, a]^T$ comprises the wheel steering angle $\delta$ (with a maximum value $\delta_m$) and the linear acceleration $a$. Applying Euler's approximation to \cref{eq:case1}, we obtain the discrete-time version of the kinematic model:
\begin{equation}
    \begin{cases}
    x_{k+1} = x_k+v_k \cos\theta_k\Delta t, \\
    y_{k+1} = y_k+v_k \sin\theta_k\Delta t, \\
    \theta_{k+1} = \theta_k+v_k\frac{tan\delta_k}{L}\Delta t,\\
    v_{k+1} = v_k+a_k\Delta t,
    \end{cases}
    \label{eq:case2}
\end{equation}
where $\Delta t$ is the sampling interval and the subscript $k$ denotes the time step. For notational simplicity, we reformulate \cref{eq:case2} as the following compact representation:
\begin{equation}
    \boldsymbol{x}_{k+1} = f_d(\boldsymbol{x}_k, \boldsymbol{u}_k).
    \label{eq:model2}
\end{equation}

Our objective is twofold:  
(1) Trajectory smoothing: to generate a smoothed trajectory represented as a sequence of key states \( \{\boldsymbol{x}_{r,n}\}_{n=0}^T\), where each state \( \boldsymbol{x}_{r,n}=[\boldsymbol{z}_n,\theta_n,v_n] \) corresponds to the vehicle’s state at time \( t_n = n\Delta t \), $\boldsymbol{z}_n=[x_n,y_n]$, and $T$ is the maximum time step. 
Note that the smoothed trajectory should start from the $\boldsymbol{r}_0$ and reach $\boldsymbol{r}_m$, with the orientation at both the start and the end consistent with that of the reference path, i.e. $\theta_0=\phi_1,\theta_T=\phi_m$, respectively. The planned trajectory must satisfy the vehicle’s kinematic constraints (see \cref{eq:case2}), avoid full-body collisions with the farmland boundaries, and minimize the deviation from the reference path \( R \). 
(2) Tracking control: to design a tracking controller that enables the vehicle to follow the planned trajectory accurately while respecting kinematic and obstacle avoidance constraints. The controller should aim to minimize tracking errors while ensuring control effort remains within physical limits. 

\section{Proposed Planning-Control Framework}
Our approach consists of two sequential stages. In the first stage, the Path-Tracking Hybrid A\* algorithm is proposed to generate a smoothed trajectory \( \{\boldsymbol{x}_{r,n}\}_{n=0}^T \). To improve adherence to the reference path, novel cost and heuristic functions are introduced, and pruning strategies are incorporated to enhance computational efficiency. In addition, a real-time obstacle avoidance extension of the Path-Tracking Hybrid A\* algorithm is developed. In the second stage, the hierarchical MPC is employed to track the planned trajectory with high accuracy and safety. Specifically, a linearized MPC is used to warm-start the nonlinear solver, thereby accelerating convergence with a minor compromise in accuracy.
\subsection{Path-Tracking Hybrid A* Algorithm}
The Path-Tracking Hybrid A* generates trajectories through a blend of discrete and continuous space exploration. Specifically, the Path-Tracking Hybrid A* discretizes the continuous search space into grids, while allowing movements based on the kinematic model of the vehicle instead of being limited to grid alignments. This approach helps the Path-Tracking Hybrid A* to closely represent vehicle dynamics and constraints, such as turning radii, in the process of finding the optimal path. 

The Path-Tracking Hybrid A* utilizes motion primitives to extend the trajectory. Specifically, as stated in \cref{eq:case1}, the motion of a vehicle at low speeds approximates circular motion with a maximum curvature constraint $c_{\text{max}} = \operatorname{tan}(\delta_m)/L$, which means the motion primitives should be straight line segments or circular arcs which have curvature less than $c_{\text{max}}$.

The search process is based on nodes. Each node $p$ contains the information of the current state $\boldsymbol{x}_p=[x_p,y_p,\theta_p]^T$, the total evaluation function $f(p)$ (introduced in \cref{deviation}), and the previous node $p_{prev}$, which is denoted as $p=[\boldsymbol{x}_p,f(p),p_{\text{prev}}]$. Note that each grid can only contain one forward and one backward node, respectively. In the following, we will introduce the evaluation function $f$ first.

\subsubsection{Deviation-Aware Cost and Heuristic for Path Smoothing}
\label{deviation}
\begin{figure}[!t]
    \centering
    \begin{minipage}{.23\textwidth}
        \centering
        \begin{itemize}
            \item \includegraphics[width=1\linewidth]{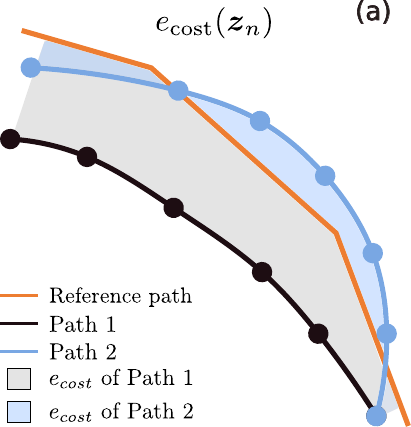}
        \end{itemize}
        
    \end{minipage}%
    \hfill
   \begin{minipage}{.23\textwidth}
        \centering
        \includegraphics[width=.85\linewidth]{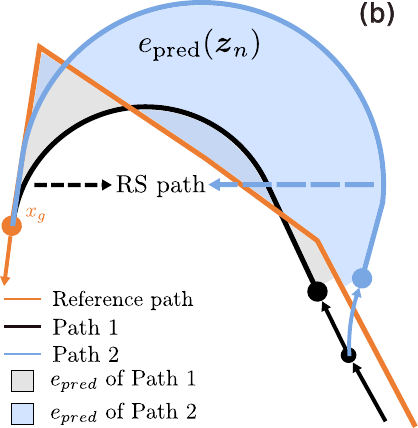}
    \end{minipage}
    \caption{Illustration for the cost (a) and heuristic (b) for path smoothing. (a): the shaded area illustrates the deviation cost calculated by \cref{eq:ecost}, and the $e_{\text{cost}}$ of Path 1 is larger than that of Path 2. \space (b): Two nodes are extended from their parent node, and each is connected to the goal point with a Reed-and-Shepp path. Note that due to the divergent orientation of Path 2, the $e_{\text{pred}}$ of Path 2 is significantly larger than that of Path 1. Therefore, the heuristic \cref{eq:epred} is useful in burning the wrong heading nodes.}
    \label{fig:1}
\end{figure}
One of the most significant contributions of our work is the design of deviation-aware cost and heuristic functions to minimize the tracking error for path smoothing. 
Newly designed formulations are employed to characterize the degree of deviation, thereby enabling the generated trajectory to more closely conform to the reference path.
First, we defined the distance from a point $\boldsymbol{z}\in\mathcal{P}$ to the polyline reference path $R$: 
\begin{equation}
    dis(\boldsymbol{z},R) = \min_{i=1,2...m} d(\boldsymbol{z},l_i),
    \label{eq:dis}
\end{equation}
where $d(\boldsymbol{z},l_i)$ denotes the distance from the point $\boldsymbol{x}$ to the line segment $l_i$. For notation simplicity, in the following discussion, we mix the notation of a node and its position, since there is almost a one-to-one correspondence. Based on \cref{eq:dis}, we propose a cost function $e_{\text{cost}}(\cdot)$ to evaluate the current deviation degree and a heuristic function $e_{\text{pred}}(\cdot)$ to estimate the future deviation degree. For the former function, we proposed a new metric of the deviation degree between the robot's generated trajectory and the reference path $R$:
\begin{equation}
    e_{\text{cost}}(\boldsymbol{z}_n) = \sum_{i=1}^{n} dis(\boldsymbol{z}_i,R)\cdot ||\boldsymbol{z}_i-\boldsymbol{z}_{i-1}||,
    \label{eq:ecost}
\end{equation}
where $\boldsymbol{z}_n$ is the position of the state $\boldsymbol{x}_{r,n}$ in the smoothed trajectory, and $||\cdot||$ is the $L_2$ norm.
This metric takes into account the cumulative deviation along the trajectory length and can approximately be regarded as the area enclosed by the generated trajectory and the reference path, which is shown in \cref{fig:1} (a). The \cref{eq:ecost} can be calculated recursively, i.e., 
\begin{equation}
    e_{\text{cost}}(\boldsymbol{z}_n)=e_{\text{cost}}(\boldsymbol{z}_{n-1})+dis(\boldsymbol{z}_n,R)\cdot ||\boldsymbol{z}_n-\boldsymbol{z}_{n-1}||.
\end{equation}

Analogous to \cref{eq:ecost}, we aim to compute the heuristic $e_{\text{pred}}(\boldsymbol{z}_n)$ as the deviation between the predicted trajectory and the reference path. Different from the  \cref{eq:ecost} above, $e_{\text{pred}}(\boldsymbol{z}_n)$ requires cumulating the deviation along the robot's predictive trajectory that remains unknown. To tackle this issue, we propose to use a Reed-and-Shepp path, which connects $(\boldsymbol{z}_n, \theta_n)$ and $(\boldsymbol{r}_m, \phi_m)$ by straight lines and circular arcs that have curvature less than $c_{\text{max}}$.  If the orientation of the current node significantly diverges from the reference path, the accumulated deviation distance along the Reed-and-Shepp (RS) path will be substantial, leading to a significantly large $e_{\text{pred}}$, which is illustrated in \cref{fig:1} (b). Thus, the heuristic function is defined:
\begin{equation}
    e_{\text{pred}}(\boldsymbol{z}_n) = \sum_{j=1}^{J}dis(\boldsymbol{s}_j,R) \cdot ||\boldsymbol{s}_j-\boldsymbol{s}_{j-1}||,
    \label{eq:epred}
\end{equation}
where $\{\boldsymbol{s}_j\}_{j=0}^J\subset P$ are key points sampled from the obtained RS path. Still, we need to employ a supplementary function to guide the trajectory towards $\boldsymbol{r}_m$, since in some cases, such as when the motion primitives are nearly perpendicular to the reference path, the deviation cost does not increase. The supplementary function can be defined as:
\begin{equation}
    h(\boldsymbol{z}_n) = \sum_{i=1}^{n} ||\boldsymbol{z}_j-\boldsymbol{z}_{j-1}||+\sum_{j=1}^{J} ||\boldsymbol{s}_j-\boldsymbol{s}_{j-1}||,
    \label{eq:h}
\end{equation}
which depicts the overall distance from start to end, including both the real and the predicted.  Minimizing $h(\boldsymbol{z}_n)$ prevents new motion primitives with divergent directions. By incorporating the above functions \cref{eq:ecost,eq:epred,eq:h}, the total evaluation function for optimization is defined as follows:
\begin{equation}
    f(\boldsymbol{z}_n) = \alpha \cdot e_{\text{cost}}(\boldsymbol{z}_n)+\beta\cdot e_{\text{pred}}(\boldsymbol{z}_n).
    \label{eq:totalcost}+\gamma\cdot h(\boldsymbol{z}_n),
\end{equation}
where $\alpha,\beta,\gamma$ are coefficients representing the weights.

\subsubsection{Algorithm Overview}
\label{path tracking}
\begin{algorithm}[!t]
    \DontPrintSemicolon
    \KwIn{ Reference path $R$, farmland $P$}
    \KwOut{Smoothed trajectory $\{\boldsymbol{x}_{r,n}\}_{n=0}^T$}
    \BlankLine
    Initialize an empty priority queue $O$\label{initialize}\;
    Initallize the start node $s\gets[(\boldsymbol{r}_0,\theta_0),0,\operatorname{Null}]$\;
    Initiallize the end node $t\gets \operatorname{Null}$\;
    $O.\operatorname{insert}(s)$ \label{insert1}\;
    \While{$O$ is not empty}{
        $p \gets O.\operatorname{pop}()$ \label{pop}\;
        $\boldsymbol{z}_p\gets p.\operatorname{position()}$\;
        \If{$\boldsymbol{z}_p$ == $\boldsymbol{r}_m$}{ \label{success}
            \Return $p.\operatorname{backtrack()}$
        }
        \If{$\operatorname{dist}(\boldsymbol{z}_p, \boldsymbol{r}_m) < d_0$}{\label{close}
        $C_{RS}\gets \operatorname{RS}(\boldsymbol{z}_p,\boldsymbol{r}_m)$\;
         \If{$\neg \operatorname{collide}(C_{RS},P)$}{
         \label{RS}
         \If{$t \neq \operatorname{Null}$ and $ f(p)>f(t)$}{
         \textbf{Continue} \tcp{Pruning 1} \label{prune1}
         }
         \If{$t \neq \operatorname{Null}$}{
         $O.\operatorname{remove}(t)$
         }
         $t\gets[(\boldsymbol{r}_m,\theta_m),f(p),p]$\;
         $O.\operatorname{insert}(t)$\;
         } 
        }
        \ElseIf{$t\neq \operatorname{Null}$}{\textbf{Continue} \tcp{Pruning 2} \label{prune2}} 
    
     $O.\operatorname{insert}(\operatorname{ValidChildren}(p, P))$
     \label{insert2}\;
     }
    \If{$O$ is empty}{ \label{fail}
     \Return $\varnothing$}
    \caption{Path Tracking Hybrid A* Algorithm}
    \label{alg:1}
\end{algorithm}

\Cref{alg:1} takes the reference path $R$ and the farmland polygon $P$ as input. 
The algorithm begins by initializing a priority queue $O$ that is used to store all the currently searching nodes.
In each loop, the node with the smallest $f$ will be popped from $O$ (\cref{pop}). When $\operatorname{dist}(\boldsymbol{z}_p,\boldsymbol{r}_m)$ is less than a specific constant $d_0$ (\cref{close}), i.e.,  the current node $p$ is in the vicinity of the target position $\boldsymbol{r}_m$, the algorithm operates an analytical expansion, which connects $\boldsymbol{z}_p$ and $\boldsymbol{r}_m$ with a RS path, denoted as $C_{RS}$ (\cref{RS}). Note that $\operatorname{dist}(\boldsymbol{z}_p,\boldsymbol{r}_m)$ is measured by the distance along the reference path between their projections on the reference path.
If \( C_{RS} \) is obstacle-free, then the predicted \( e_{\text{pred}}(p) \) and the corresponding portion of the \( h(p) \) based on the RS path are feasible. Consequently, the target node at \( \boldsymbol{r}_m \) will be assigned the same evaluation function value as \( f(p) \). Note that at any given time, at most one node is maintained at \( \boldsymbol{r}_m \). If there is no existing target node \( t \) at \( \boldsymbol{r}_m \), a new node extended to $\boldsymbol{r}_m$ from \( p \) is inserted into the open set \( O \). If a target node \( t \) already exists, its evaluation function value \( f(t) \) is compared with \( f(p) \); if \( f(p) < f(t) \), the existing node \( t \) is replaced with the new one. Otherwise, the node \( p \) is discarded.

In \cref{insert2}, starting from node \( p \), the expansion to child nodes is performed using motion primitives. These motion primitives involve discretized curvature values within the range \( [-c_{\text{max}}, c_{\text{max}}] \), which include straight-line segments (curvature \( 0 \)) and both forward and reverse directions. Each motion primitive is applied to simulate the robot’s motion using a kinematic model over a fixed step length, yielding a set of candidate child nodes.
For each generated child node, a validity check is conducted to ensure no collision between the robot’s body and the environmental boundaries. If the node is valid, its evaluation cost is computed and it is inserted into the open set 
\( O \) for further exploration. Nodes that fail the validity check or lead to collisions are discarded, thereby ensuring full-body collision avoidance along the generated trajectory.

Finally, if the popped node $p$ is located at the end point $\boldsymbol{r}_m$ (\cref{success}), the desired path can be found by iteratively tracing back to each preceding node. If the priority queue is empty before the path is found, as shown in \cref{fail}, the algorithm will return $\varnothing$, i.e., path not found.

By applying the aforementioned method, we obtained the smoothed trajectory $\{\boldsymbol{x}_{r,n}\}_{n=0}^T$. For each $\boldsymbol{x}_{r,n}$, both position and orientation components have already been determined. To construct the complete state required for the subsequent MPC tracking stage, we augment each $\boldsymbol{x}_{r,n}$ with a velocity component, assigning a constant reference speed $v_r$.

\subsection{Pruning Techniques and Real-time Replanning}
In this section, we first present pruning techniques that substantially reduce the computational burden, followed by the introduction of an online replanning strategy designed to facilitate real-time avoidance of unforeseen obstacles during operation.

\subsubsection{Pruning Techniques}
\label{subsubsec:pruning}
A too-large priority queue can lead to excessively long search times, calling for effective pruning strategies. Two pruning strategies are proposed as follows:

\textbf{1.} In the vicinity of the target node \( \boldsymbol{r}_m \), branches with low potential are pruned. The first pruning technique (\cref{prune1}) dictates that if a node \( t \) already exists in the open set \( O \) and has a lower evaluation function than the current node \( p \), then the present node \( p \) is discarded. In other words, the process of generating child nodes, as described in \cref{insert2}, is not carried out, thus reducing further branches. Since nodes near \( \boldsymbol{r}_m \) require time-consuming analytical expansions using the RS path (\cref{RS}), this pruning technique effectively eliminates branches close to \( \boldsymbol{r}_m \) early on by evaluating their potential to have a lower evaluation function, which significantly accelerates the overall search process.

\textbf{2.} If a target node \( t \) is already present in the priority queue \( O \), distant branches are pruned. The second pruning strategy, described in \cref{prune2}, ensures that if there is at least one valid target node in \( O \), and the current node \( p \) is not in the vicinity of \( \boldsymbol{r}_m \), then node \( p \) is discarded. If a valid target node \( t \) already exists, a satisfactory result has been achieved. This strategy prevents additional nodes from entering the neighborhood of \( t \), thereby significantly reducing the computational effort required for further analytical expansions.

\subsubsection{Replanning Strategy for Real-time Obstacle Avoidance}
\label{real-time-theory}
Although the smoothed trajectories can be generated in advance by methods in \cref{path tracking}, unexpected obstacles, such as other agricultural vehicles parked on the farmland, could occur. Therefore, real-time and safe replanning of the smoothed trajectory plays a vital role in the practical execution of the planning-control algorithm. 

The real-time obstacle avoidance problem is formulated as follows. The set of obstacles is defined as $\mathcal{O}_b = \{O_b | O_b \subset P\}$, in which $O_b$ is a convex polygon defined as $O_b = \{o_1, o_2, \dots, o_{n_b}\}$, where $o_i \in \mathbb{R}^2, i=1,2,\dots,n_b$ are the vertices and $n_b$ is the number of vertices. Note that the obstacles are not known before they are detected by the vehicle during driving. The vehicle has a limited detection range of $R_f\in\mathbb{R}$ and a limited field of view (FOV) $\alpha_f\in[0,\pi]$, which forms a fan-shaped area $F_f\in\mathbb{R}$. When all the vertices of $O_b$ have been in the area of $F_f$, the obstacle is detected by the vehicle. In this circumstance, we need to replan the trajectory to ensure the collision-free driving of the vehicle.

The working principle of the Path-Tracking Hybrid A* algorithm, when the vehicle encounters random obstacles, is as follows: First, the vehicle identifies the start and end states along the original smoothed trajectory that are in collision with the obstacles. Subsequently, the range of the original trajectory to be replanned is expanded by a few points on either side. This adjusted range serves as the reference trajectory for real-time replanning using the Path-Tracking Hybrid A* algorithm, which generates a new trajectory that ensures full vehicle collision avoidance. Finally, the replanned trajectory replaces the corresponding portion of the original trajectory, which is then tracked using our hierarchical MPC controller, as detailed in \cref{MPC}.

Several details of trajectory planning are different from \cref{alg:1}. First, the triggering condition $\operatorname{dist}(\boldsymbol{z}_p, \boldsymbol{r}_m) < d_0$ in \cref{close} is changed to $\neg \operatorname{intersect}(\boldsymbol{z}_p, \boldsymbol{r}_m, \mathcal{O}_b)$, which means node $p$ connects to the end point $\boldsymbol{r}_m$ with a straight line without intersection with obstacles. This means that only when a node has already bypassed the obstacles will it carry out analytic expansion to the end point. 
Second, we propose an iterative method for collision checking to enhance the replanning speed. Specifically, during the replanning process, we inflate both the obstacles and boundary edges by a certain width. Collision checking is initially performed only at the vehicle's rear axle center in each planning node. Once a new trajectory is generated, full vehicle collision checking is conducted along the entire path. If any collisions with the vehicle occur, the collided edges, either from the obstacles or the boundary, are progressively inflated, and the replanning process is rerun. This iterative procedure continues until a collision-free trajectory is obtained.

By leveraging the pruning techniques in \cref{subsubsec:pruning} and the real-time replanning strategy in \cref{real-time-theory}, our algorithm is capable of effectively avoiding unforeseen obstacles.

\subsection{Hierarchical MPC-based Trajectory Tracking}
\label{MPC}
This section proposes a hierarchical MPC method to safely track the smoothed trajectory $\{\boldsymbol{x}_{r,n}\}_{n=0}^T$ generated in \cref{alg:1}, which serves as a reference trajectory in the MPC framework. The method is divided into two stages. First, an initial control is generated using linearized MPC with linear constraints to obtain an optimal solution in a shorter time frame. Then, using this control as the initial value, local adjustments are made through nonlinear optimization with nonlinear, strict constraints, ensuring that the solution accurately satisfies the real-world constraints.

First, the linearized MPC method utilizes the system dynamics to predict future vehicle behavior and generate an optimal control sequence to minimize the tracking error, which is formulated as follows:

\begin{subequations}
    \begin{align}
\min_{\boldsymbol{U}_k} \ \ \ &\Phi_k(\boldsymbol{X}_k, \boldsymbol{U}_k) \label{opt0}\\
  s.t. \quad & \cref{eq:linearmodel},  \\
  & C_k\boldsymbol{X}_k<\boldsymbol{d}_k, \label{eq:constraint0} \\
  & |a_{k+j}| \leq a_m,\quad j=0,\dots,N-1,  \label{eq:accel}\\
  & |\delta_{k+j}| \leq \delta_m,\quad j=0,\dots,N-1, \label{eq:angle}
\end{align}
\end{subequations}
where $a_m$ in \cref{eq:accel} denotes the maximum acceleration,  $\delta_m$ in \cref{eq:angle} denotes the maximum steering angle of the vehicle, and \cref{eq:constraint0} denotes the linear collision avoidance constraint between the agriculture robot and the field, specified by $C_k$ and $\boldsymbol{d}_k$. Vectors $\boldsymbol{X}_k$ and $\boldsymbol{U}_k$ are defined as the concatenation of state and control vectors $\boldsymbol{x}$ and $\boldsymbol{u}$ respectively, specified as:
\begin{equation}
    \boldsymbol{X}_k = \begin{bmatrix}
\boldsymbol{x}_{k+1} \\
\boldsymbol{x}_{k+2} \\
\vdots \\
\boldsymbol{x}_{k+N}
\end{bmatrix}
,\boldsymbol{U}_{k} = \begin{bmatrix}
\boldsymbol{u}_{k} \\
\boldsymbol{u}_{k+1} \\
\vdots \\
\boldsymbol{u}_{k+N-1}
\end{bmatrix}
. \label{eq:concatenated states}
\end{equation} 
For the sake of brevity, we define the error vectors of the robot state and control input 
as $\tilde{\boldsymbol{x}}_{k}=\boldsymbol{x}_{k}-\boldsymbol{x}_{r,k}$, $\tilde{\boldsymbol{u}}_{k}=\boldsymbol{u}_{k}-\boldsymbol{u}_{r,k}$, respectively, where  $\boldsymbol{u}_{r,k}$ denotes the control vectors of the reference trajectory to be tracked. Note that the control vector $\boldsymbol{u}_{r,k}$ of the reference trajectory is set to, $[\operatorname{arctan}(c_kL),0]^T$, where $c_k$ is the curvature of the $k^{\text{th}}$ motion primitive in the reference trajectory. Analogous to \cref{eq:concatenated states}, we define $\tilde{\boldsymbol{X}}_k$, $\boldsymbol{X}_{r,k}$ and $\tilde{\boldsymbol{U}}_k$ as the concatenation vectors of state error $\tilde{\boldsymbol{x}}$, reference trajectory $\boldsymbol{x}_r$ and control error $\tilde{\boldsymbol{u}}$ in the predictive horizon, respectively.
The objective function to be minimized in \cref{opt0} is proposed as follows:
\begin{equation}
    \Phi_k(\boldsymbol{X}_k, \boldsymbol{U}_k)  = \sum_{j=1}^N \Bigl( \tilde{\boldsymbol{x}}^T_{k+j} Q \tilde{\boldsymbol{x}}_{k+j} + \label{opt} 
    \tilde{\boldsymbol{u}}^T_{k+j-1}  W \tilde{\boldsymbol{u}}_{k+j-1} \Bigr),       
\end{equation}
where \( N \) is the prediction horizon and \( Q \succeq 0 \), \( W \succeq 0 \) are weighting matrices for the error in the state and control variables, and ``$\succeq$'' means semi-definiteness.
By linearizing \cref{eq:model2}, we obtain the state propagation formula as follows:
\begin{equation}
    \tilde{\boldsymbol{x}}_{k + 1} = A_{k}\tilde{\boldsymbol{x}}_{k} + B_{k}\tilde{\boldsymbol{u}}_{k},
    \label{eq:linearmodel}
\end{equation}
where:
\begin{equation}
A_{k} = \begin{bmatrix}
1 & 0 & -v_{r,k}\sin\theta_{r,k}\Delta t & \cos\theta_{r,k}\Delta t \\
0 & 1 & v_{r,k}\cos\theta_{r,k}\Delta t & \sin\theta_{r,k}\Delta t \\
0 & 0 & 1 & \frac{\Delta t}{L}\tan\delta_{r,k} \\
0 & 0 & 0 & 1
\end{bmatrix}
,
\end{equation}
\begin{equation}
B_k = \Delta t \begin{bmatrix}
0 & 0 \\
0 & 0 \\
0 & \frac{v_{r,k}}{L \cos^2\delta_{r,k}} \\
1 & 0
\end{bmatrix}
.
\end{equation}

Then, $\tilde{\boldsymbol{X}}_k$ can be expressed as a linear combination of $\tilde{\boldsymbol{x}}_k$ and $\boldsymbol{U}_k$, shown as the following:
\begin{gather}
\tilde{\boldsymbol{X}}_k = \bar{A}_k\tilde{\boldsymbol{x}}_k+\bar{B}_k\boldsymbol{U}_k, \label{linear} 
\end{gather}
\begin{figure}[!t]
    \centering
    \includegraphics[width = \linewidth]{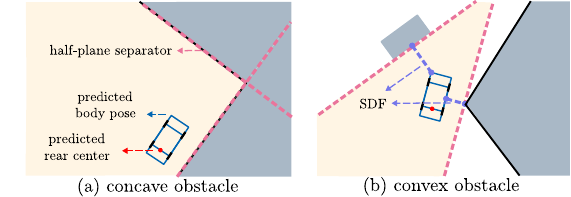}
   \caption{Illustration of how to determine the half-plane separators in the predicted states. (a) If the boundary obstacle is concave, both lines of boundary segments are used as half-plane separators. (b) If the boundary obstacle or the random obstacle is convex, the half-plane separator is determined according to SDF. The feasible region is illustrated in yellow.}
   \label{mpc_new}
\end{figure}
where $\bar{A}_k$ and $\bar{B}_k$ are coefficient matrice calculated from $A_k$ and $B_k$, detailed in \cite{kunhe2005mobile}.

 We now focus on specifying the collision avoidance constraint \cref{eq:constraint0} to enhance the robot's safety while tracking the trajectory generated in \cref{alg:1}. 
The linear constraints contain both the constraints of farmland boundaries and the constraints of random obstacles. First, given a predicted state vector $\boldsymbol{x}_{k+j}$ in the predicted horizon, we determine the boundary constraints in the following way, which is shown in \cref{mpc_new}: 
first we identify the current field boundary segment $e_i$ (the boundary segment with minimum distance to the rear center of the vehicle) and the next boundary segment along the vehicle's forward direction $e_{i+1}$, and determine whether the boundary obstacle formed by these two segments is convex or concave. 
If the boundary obstacle is concave (\cref{mpc_new}(a)), the lines of the $e_i$ and $e_{i+1}$ are both used as half-plane separators. If the boundary obstacle is convex (\cref{mpc_new}(b)), we 
use the Gilbert-Johnson-Keerthi (GJK) algorithm \cite{gilbert1988fast} or the Expanding Polytope Algorithm (EPA) \cite{van2001proximity} to find the signed distance function (SDF) $\hat{d}$ between the rectangular vehicle body and the triangle formed by segments $e_i$, $e_{i+1}$, and the line connecting their non-shared endpoints, along with the nearest points $P_1$ and $P_2$ on the vehicle and the obstacle respectively. Then, the normal vector of the half-plane separator is calculated by $\hat{n} = sgn(\hat{d})\cdot(P_1-P_2)/||P_1-P_2||$ \cite{gao2024probabilistic}. Second, as for the random convex obstacles, the method to determine their linear constraints is the same as in the convex case (\cref{mpc_new}(b)). \\  

For a certain half-plane separator $ax+by+c=0$ with constant $c$ and normal vector $(a,b)$ pointing towards the interior of the field, we can obtain the safety condition of the right front corner of the vehicle as:
\begin{align}
    ax+by&+a(l\cos\theta+w\sin\theta)\nonumber\\ 
    &+b(l\sin\theta-w\cos\theta)+c>0,
    \label{sanjiao}
\end{align}
where $l$ is the length from the rear axis to the car front, and $w$ is half the width of the vehicle. From \cref{linear} and $\tilde{\boldsymbol{X}}_k=\boldsymbol{X}_k-\boldsymbol{X}_{r,k}$, we know that:
\begin{equation}
    \boldsymbol{X}_k = \bar{A}_k\tilde{\boldsymbol{x}}_k+\bar{B}_k\boldsymbol{U}_k + \boldsymbol{X}_{r,k}.\label{eq:linearx}
\end{equation}
Note that $\boldsymbol{x}_{k+j}$ is obtained from \cref{eq:linearx} using $\tilde{\boldsymbol{x}}_k$ and the control variable $\boldsymbol{U}_k$, and thus 
$\boldsymbol{x}_{k+j}$ is an unknown priori. and the nearest boundary segment is estimated using the reference trajectory $\boldsymbol{x}_{r,k+j}$ (generally written as $[x_r, y_r, \theta_r, v_r]^T$). As we want to change \cref{sanjiao} into a linear inequality for $x,y$ and $\theta$, we linearly expand $sin\theta$ and $cos\theta$ at $\theta_r$. After simplification, the inequality can be written as:
\begin{gather}
\bar{C}_{k+j}\boldsymbol{x}_{k+j}<d_{k+j}, \label{matrix} \\
\bar{C}_{k+j}=\mathrm{diag}[a,b,c_3,0], \\
c_3 = a \left(w \cos\theta_r - l \sin\theta_r\right) + b \left(w \sin\theta_r + l \cos\theta_r\right),  \\
d_{k+j} =-a (w (\sin\theta_r - \theta_r \cos\theta_r) + l (\cos\theta_r +  \theta_r \sin\theta_r))- \nonumber  \\
        b (-w(\cos\theta_r + \theta_r\sin\theta_r) +  l (\sin\theta_r - \theta_r\cos\theta_r)) - c, 
\end{gather}
where $\bar{C}_{k+j}$ denotes the coefficient matrix and $d_{k+j}$ denotes a constant, both taking value at instant $k+j$ and predicted at instant $k$. By combining \cref{matrix} and \cref{eq:linearx}, we can obtain the linear inequality of safety constraint for the right front of the car:
\begin{gather}
C_k\boldsymbol{X}_k=C_k\left(\bar{A}_k\tilde{\boldsymbol{x}}_k+\bar{B}_k\boldsymbol{U}_k + \boldsymbol{X}_{r,k}\right) \nonumber\\
< [d_{k+1}\enspace d_{k+2}\enspace \cdots \enspace d_{k+N}]^T , \label{constraint}\\
C_k=\mathrm{diag}[\bar{C}_{k+1},\bar{C}_{k+2},\cdots,\bar{C}_{k+N}].
\end{gather}
Similarly, we can obtain the inequalities of safety constraints for the other three corners of the car, thus specifying the whole collision avoidance constraint \cref{eq:constraint0}. 
By this stage, \cref{opt} could be solved as a constrained quadratic programming problem, which is detailed in \cite{kunhe2005mobile}.

Although the predicted states of the linearized MPC are feasible, there exists linearization errors in the predicted states, leading to risks of collisions in real states. Therefore, a local adjustment using nonlinear optimization is introduced:
\begin{subequations}
    \begin{align}
\min_{\boldsymbol{U}_k'} \ \ & \sum_{j=1}^{N^{\prime}} \Bigl( \tilde{\boldsymbol{x}}^T_{k+j} Q \tilde{\boldsymbol{x}}_{k+j} + 
    \tilde{\boldsymbol{u}}^T_{k+j-1}  W \tilde{\boldsymbol{u}}_{k+j-1} \Bigr), \label{eq:opt2} \\
  s.t. \quad & g(\boldsymbol{x}_{k+j})>0,\quad j=1,2,\dots,N',\label{eq:g}\\ 
  & \cref{eq:model2}, \label{eq:kinemetic} 
\end{align}
\end{subequations}
where $N^{\prime} \leq N$ is a new prediction horizon, and $g(\cdot)$ is a function for collision determination. Specifically, $g(\boldsymbol{x}_{k+j})$ takes the position and orientation components of the state $\boldsymbol{x}_{k+j}$ to reconstruct the vehicle’s footprint, and then computes the minimum SDF to all obstacles using the aforementioned method. A positive value of $g(\boldsymbol{x}_{k+j})$ indicates that the vehicle is collision-free at that state. By the inclusion of \cref{eq:g,eq:kinemetic}, this method simulates the real motion and constraints, thereby providing a more accurate representation of the system's behavior and enhancing the reliability of the optimization process.

The meaning of the hierarchical framework is two-fold. On the one hand, nonlinear optimization is sensitive to the initial values, and a poor choice of initial guess may lead to unexpected local minima or excessive computational time. Therefore, the initial control variables obtained from the linearized MPC provide a suitable starting point, as they are already close to the true state-control variables, thereby offering both computational efficiency and accuracy. On the other hand, the reference trajectories generated by the Path-Tracking Hybrid A* algorithm are guaranteed to be collision-free at the vehicle's scale. Consequently, using these reference trajectories as the basis for optimization enhances safety. By solving \cref{eq:opt2}, we obtain the final control variable
$\boldsymbol{u}_k$.
\section{Simulation Results and Analysis}
\label{simulation}
\begin{figure}[!t]
    \centering
    \includegraphics[width = \linewidth]{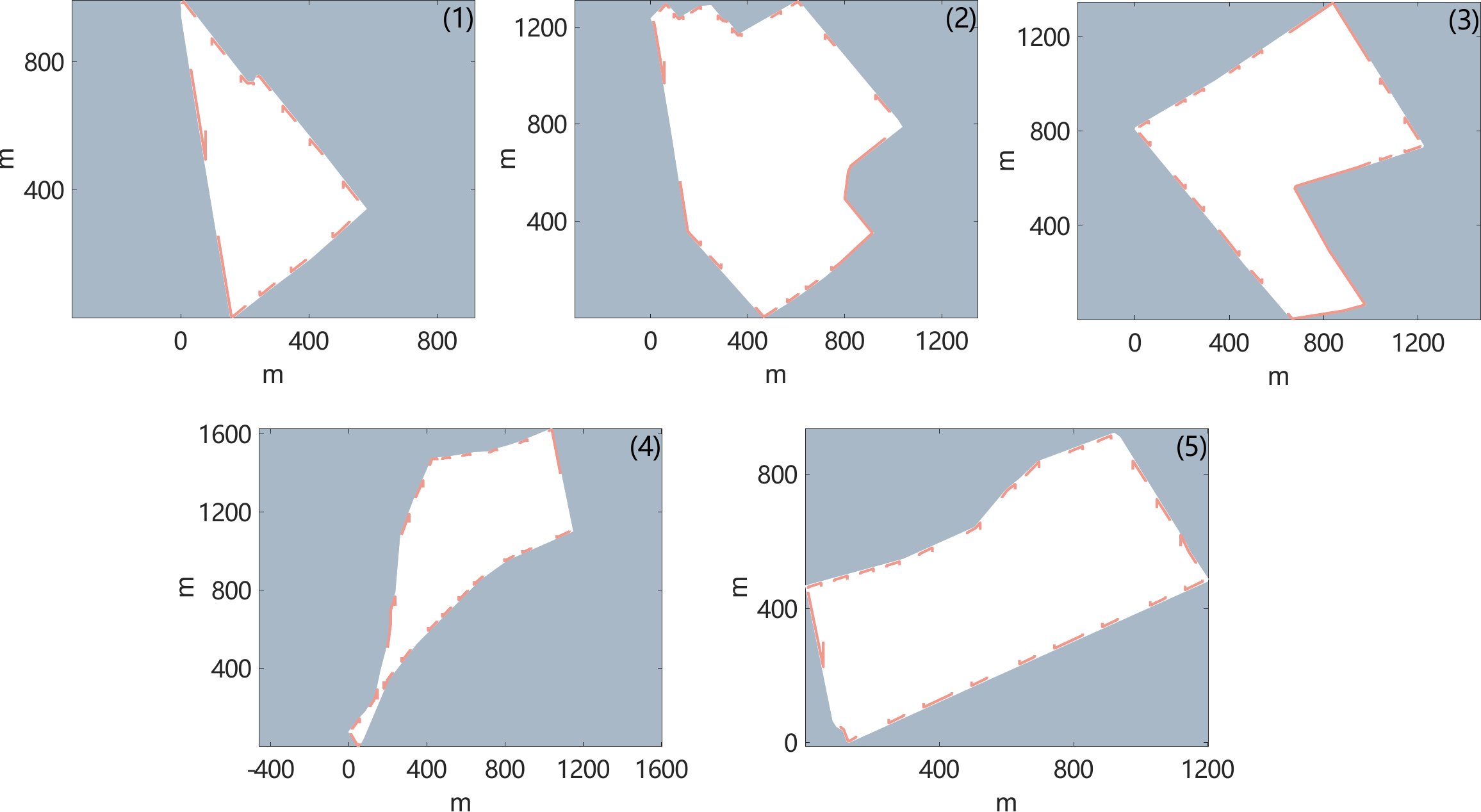}
   \caption{The five scenarios of farmland and reference path illustrations. Note that we only present 88 reference paths within a total of 713 paths.}
   \label{scenario}
\end{figure}
\begin{figure}[!t]
    \centering
    \includegraphics[width = \linewidth]{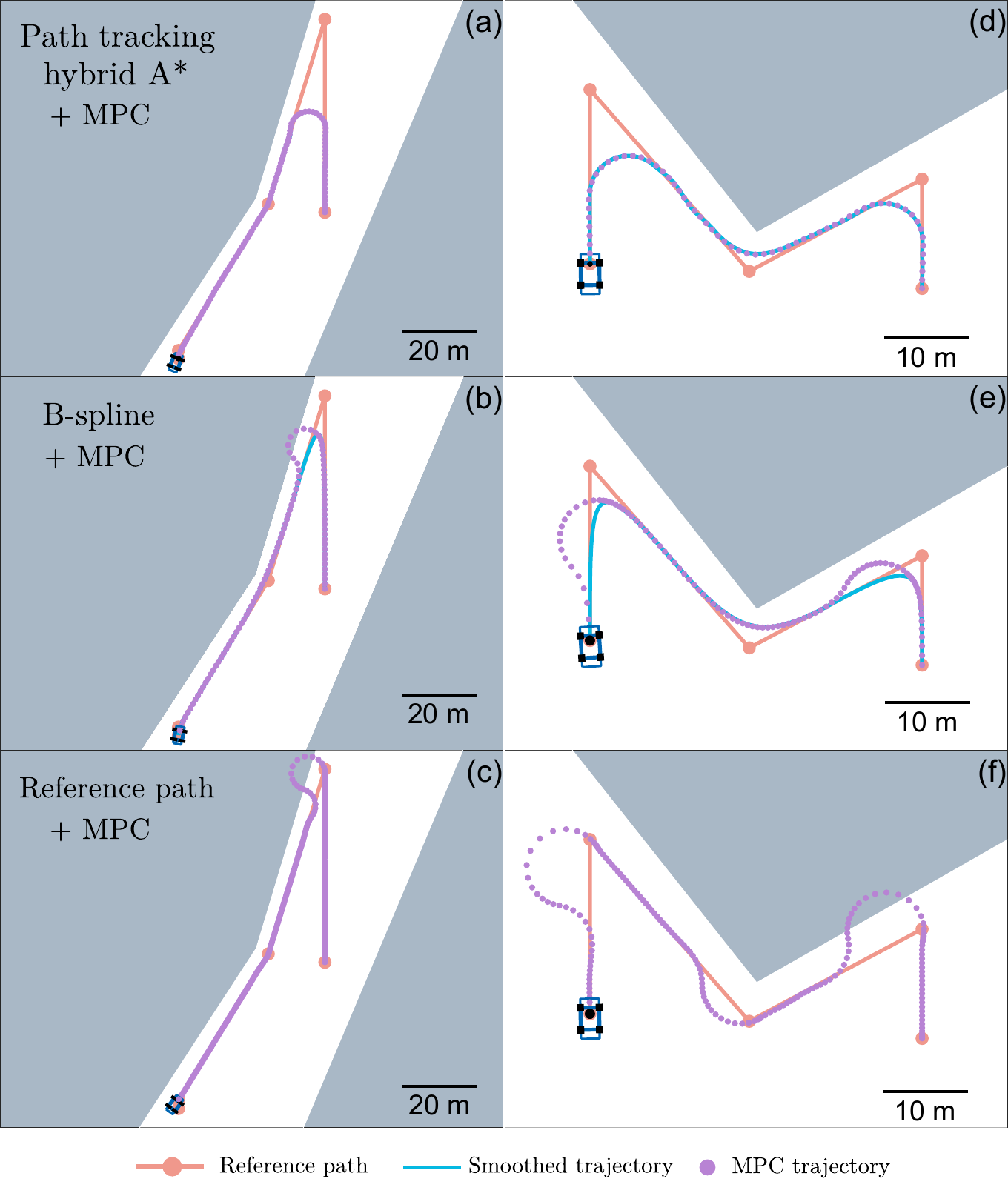}
    \caption{Comparison of three different methods on two typical scenarios. Note that each row represents experiments of the same method in different scenarios. }
    \label{comparison}
\end{figure}
To evaluate the proposed algorithm, our simulation is divided into three parts. First, we assess the path-smoothing performance of the proposed Path-Tracking A* algorithm (\cref{alg:1}) in comparison with the representative collision-free B-spline path smoothing method \cite{noreen2020collision}. Second, by using the MPC technique introduced in \cref{MPC}, we simulate the whole trajectory planning-control system and evaluate the success ratio and real deviation degree of MPC trajectories, thereby assessing the tracking feasibility of the smoothed trajectory and the performance of the hierarchical MPC. Third, we evaluate the planning-control system under scenarios with randomly generated obstacles to examine its obstacle avoidance capability.

Simulations in five farmland scenarios are conducted, along with a total of 713 reference paths located on these farmlands, as shown in \cref{scenario}. We should point out that the farmland and reference paths are all based on real-world data. 
As for the vehicle parameters, the length is \qty{4.7}{m}, the width is $\qty{2.2}{m}$, the wheelbase $L$ is $\qty{2.6}{m}$, and the maximum steering angle $\delta_m$ is $30^\circ$. As for the Path-Tracking A* parameters, the number of discretized curvature is set to 5, the speed of the vehicle $v_r$ is set to $\qty{2}{m/s}$, and the interval of each expansion $\Delta t$ is $\qty{0.5}{s}$, and when encountering longer reference trajectories, we divide them into multiple shorter paths to avoid having too many points in priority queue $O$. As for the B-spline parameters, the order of clamped B-splines is set at 3, and the midpoints of each segment are added to the reference points before generating the B-spline, since directly employing the B-spline method may yield poor performance due to sparse waypoints of the reference path. Note that here we prioritize ensuring the smoothed path is collision-free. Therefore, methods such as \cite{elbanhawi2015continuous} that achieve B-spline curvature constraints by adjusting control points are not applicable in this context. The experiments are implemented in MATLAB R2023a (MathWorks, Inc.) on a laptop with an AMD Ryzen 7 5800H processor.
\subsection{Evaluating Path Smoothing}
\begin{figure}[!t]
    \centering
    \includegraphics[width=\linewidth]{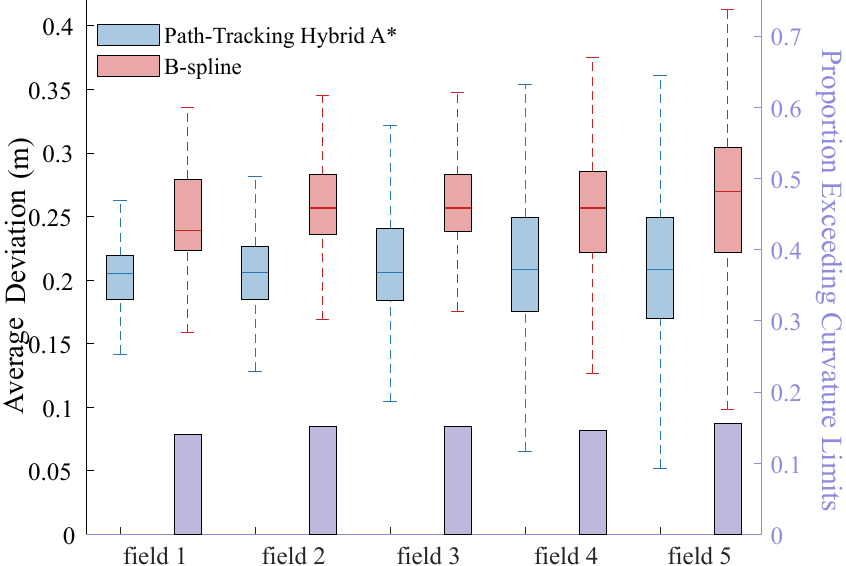}
    \caption{Average deviation degree between path-tracking Hybrid A* and the B-spline (depicted in the central box chart), and the proportion of B-spline exceeding curvature limits (shown in the lower purple bar chart).}
    \label{fig:deviation1}
\end{figure}
We now focus on quantitatively evaluating the smoothed trajectories generated by Path-Tracking Hybrid A* from two aspects, the closeness to the reference trajectory and the offset ratio of the curvature. First, we introduce the \textit{average deviation degree} as a major closeness metric formulated as follows:
\begin{equation}
    E(\{\boldsymbol{x}_{r,n}\}_{n=0}^T) = \frac{\sum_{i=1}^{T} ||\boldsymbol{z}_i-\boldsymbol{z}_{i-1}||\cdot dis(\boldsymbol{z}_i,R)}{\sum_{i=1}^{T} ||\boldsymbol{z}_i-\boldsymbol{z}_{i-1}||},
    \label{evaluation}
\end{equation}
which is calculated by dividing the total deviation as defined in \cref{eq:ecost} by the total trajectory length, representing the average deviation degree.

Implementing our path tracking Hybrid A* method and the B-spline method  \cite{noreen2020collision}, we get the statistical results of the average deviation degree and the proportion of sample points that exceeded the curvature constraints as shown in the \cref{fig:deviation1}. As seen in the box chart in \cref{fig:deviation1}, our algorithm outperforms the B-spline method in terms of the average deviation degree in all field scenes, owing to its explicit incorporation of the deviation degree into the cost function.
Additionally, the proportion of B-spline trajectories exceeding the curvature bound approaches 20\%, while that of the Path-Tracking Hybrid A* remains at 0\%, highlighting our method’s consistent adherence to curvature constraints. This is due to the use of motion primitives that inherently respect these limits. Overall, our smoothed trajectory stays closer to the reference path and better satisfies curvature constraints, enabling more reliable tracking by the backend MPC, as further discussed in the next subsection.

\subsection{Evaluating the MPC Trajectory Tracking}
\begin{figure}[!t]
    \centering
    \includegraphics[width=\linewidth]{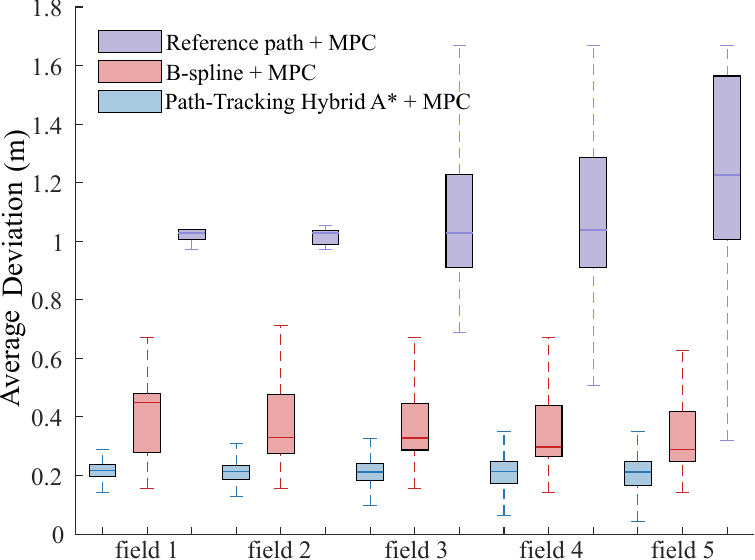}
    \caption{Average deviation degree of three different MPC trajectories.}
    \label{fig:boxchart2}
    \includegraphics[width = \linewidth]{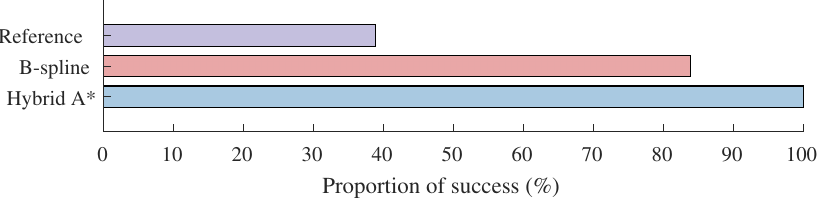}
    \caption{Proportion of success of three different MPC trajectories.}
    \label{bar2}
\end{figure}
\begin{figure*}[t]
    \centering
    \includegraphics[width=\linewidth]{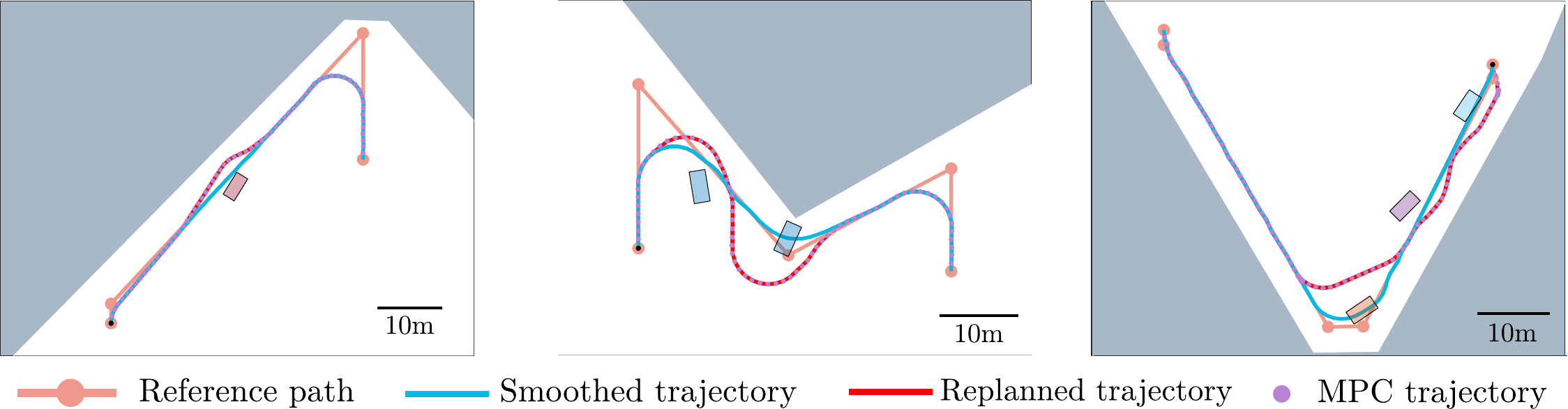}
    \caption{Typical results of the real-time obstacle avoidance of the planning-control system, under the circumstances of 1, 2, and 3 random obstacles, respectively.}
    \label{fig:showcase}
\end{figure*}
In this subsection, we use the hierarchical MPC approach introduced in \cref{MPC} to track three kinds of paths: paths smoothed by the proposed Path-Tracking Hybrid A*, paths smoothed by the collision-free B-spline method\cite{noreen2020collision}, and the raw reference path. To enable a fair comparison with the baseline, we make adaptations in the following two aspects. First, since the smoothed trajectories generated by the baseline methods fail to account for potential body collisions, they often violate safety constraints, making it difficult for the backend MPC to converge to a feasible solution. As a result, local nonlinear adjustment is not performed in cases where baseline methods are used. Also, as the smoothed trajectories generated by baseline methods yield poor tracking feasibility, rendering the optimization problem \cref{opt0} unsolvable in some cases, we adopt a unified method in baseline cases: when an unsolvable situation occurs, we remove the constraint \cref{eq:constraint0} and implement a collision detection after each track to evaluate the safety of the robot trajectory. We set the prediction horizon $N$ of MPC at 20, and we define failure as occurring under the following conditions: when the vehicle collides with obstacles, or when the vehicle ends up with a distance greater than \qty{10}{m} from the endpoint or with an angle greater than $60^\circ$ relative to the end point $\boldsymbol{r}_m$.

The tracking trajectories in representative scenarios are shown in \cref{comparison}. For the raw reference path, the vehicle encounters sharp turns with theoretically infinite curvature at certain points, rendering it difficult for the MPC to accurately track the path. Consequently, the vehicle deviates significantly from the reference, resulting in failure, as illustrated in \cref{comparison}.
In the B-spline case, although the curvature at turning points is reduced compared to the raw path, it still exceeds the vehicle’s curvature limits at several locations. This leads to moderate deviations and occasional tracking failures.
In contrast, our method produces trajectories that remain close to the reference path while strictly satisfying the vehicle's nonholonomic and curvature constraints. As a result, these trajectories can be smoothly and safely tracked by the MPC without collision.

The deviation degree and the proportion of successful tracking are presented in \cref{fig:boxchart2} and \cref{bar2}, respectively. As shown in \cref{fig:boxchart2}, our path-tracking Hybrid A* smoothing algorithm achieves significantly better adherence to the reference path in simulation. Meanwhile, \cref{bar2} indicates that our method achieves a 100\% tracking success rate across all tested reference paths, highlighting its safety and reliability. 
In contrast, both the B-spline and raw reference path result in larger deviations and lower tracking success rates when used with MPC. Additionally, the computational efficiency of the proposed hierarchical MPC is demonstrated by its average computation time of 0.0366 seconds and variance of \(8.7 \times 10^{-5}\), indicating its suitability for real-time applications.

\subsection{Test on Real-Time Obstacle Avoidance} 
\label{real-time}
\begin{table}[!t]
\centering
\caption{Statistical results of real-time obstacle avoidance}
\label{table: real_time}
\begin{tabular}{|c|c|c|c|c|}
\hline
\multicolumn{2}{|c|}{\textbf{number of obstacles}}  & 1 & 2 & 3 \\ 
\hline
\multicolumn{2}{|c|}{\textbf{success ratio}} & 100\% & 98.8\% & 97.4\%\\
\hline
\multirow{2}{*}{\makecell[c]{\textbf{replanning} \\ \textbf{time (s)}}} & mean  & 0.6449 & 0.6713 & 0.7782 \\ \cline{2-5}
 & variance & 0.2639 & 0.4138 & 0.7668 \\ \hline
\multirow{2}{*}{\makecell[c]{\textbf{control} \\ \textbf{time (s)}}} & mean  & 0.0518 & 0.0733 & 0.0990 \\ \cline{2-5}
 &  variance & 5.0e-4 & 0.0011 & 0.0024 \\ \hline
\end{tabular}
\end{table}

This section exhibits the test results of the real-time obstacle avoidance performance of the planning-control system. To carry out the experiments, we chose 10 typical reference paths, and randomly set rectangular obstacles around the reference path. The width and height of the obstacles are set at \qty{2}{m} and \qty{4}{m}, which is almost the size of an agricultural vehicle. The detection range of the vehicle is set at \qty{15}{m}, and the FOV is set at $90^\circ$. For each scenario, we fix the number of obstacles and conduct 50 trials with randomly generated obstacle configurations. 
Note that occasions where the obstacles block the start pose or the end pose are excluded. Some typical trajectories are shown in \cref{fig:showcase}, and the statistical results are shown in \cref{table: real_time}. 

As illustrated in \cref{fig:showcase}, the trajectories replanned by the path-tracking Hybrid A* remain close to the original smoothed trajectories and can be seamlessly integrated with the remaining segments. Moreover, the hierarchical MPC enables the vehicle to safely track the replanned trajectories, even when navigating through narrow passages.
According to \cref{table: real_time}, as the number of obstacles increases, the obstacle density rises and feasible paths become more constrained, leading to gradual increases in failure rate, planning time, and control time. Nevertheless, under low obstacle densities (1–3 obstacles), the proposed planning-control framework consistently demonstrates high success rates and low computational times for both planning and control.
\section{Conclusion}
We propose a novel cross-furrow path planning and control framework for agricultural vehicles, integrating a path-tracking Hybrid A* planner with a coupled hierarchical MPC controller. To minimize deviation from the reference trajectory while respecting strict curvature and dynamic constraints, we devise novel cost and heuristic functions in the planner and incorporate linearized warm-starting in the MPC for improved convergence. Furthermore, we propose an online replanning strategy as an extension that enables real-time avoidance of unforeseen obstacles along with pruning techniques that enhances computational efficiency.
The proposed framework enables the generation of smooth, obstacle-free, and dynamically feasible trajectories for agricultural vehicles, supporting both offline smoothing and online planning, as demonstrated in experiments on real-world farm datasets. Compared to baseline methods, our approach achieves superior performance in path adherence, safety, computational efficiency, and real-time responsiveness.
Future work includes extending the framework to full-field path planning for maximizing field coverage while minimizing deviation, and exploring multi-vehicle collaboration strategies to optimally allocate agricultural tasks while maintaining global constraint satisfaction.

\bibliography{ref}
\bibliographystyle{ieeetr}
\end{document}